\documentclass[10pt,journal,compsoc]{IEEEtran}
\usepackage{color}
\usepackage{overpic}
\usepackage{amsmath}
\usepackage{multirow}
\usepackage{array}
\usepackage{booktabs} 
\usepackage{url}

\ifCLASSOPTIONcompsoc
  \usepackage[nocompress]{cite}
\else
  \usepackage{cite}
\fi

\ifCLASSINFOpdf
 
\else
 
\fi

\begin{document}
%
\title{Fairness in Deep Learning:\\ A Computational Perspective}
%
%
%
%

\author{Mengnan Du, Fan Yang, Na Zou, Xia Hu
\IEEEcompsocitemizethanks{\IEEEcompsocthanksitem Mengnan Du, Fan Yang, Xia Hu are with the Department
of Computer Science and Engineering, Texas A\&M University. 
\IEEEcompsocthanksitem Na Zou is with the Department of Industrial and Systems Engineering, Texas A\&M University.
\IEEEcompsocthanksitem E-mail: \{dumengnan,nacoyang,nzou1,xiahu\}@tamu.edu
}
}

\IEEEtitleabstractindextext{%
\begin{abstract}
Deep learning is increasingly being used in high-stake decision making applications that affect individual lives. However, deep learning models might exhibit algorithmic discrimination behaviors with respect to protected groups, potentially posing negative impacts on individuals and society. Therefore, fairness in deep learning has attracted tremendous attention recently. 
We provide a 
review covering recent progresses to tackle algorithmic fairness problems of deep learning from the computational perspective. Specifically, we show that interpretability can serve as a useful ingredient to diagnose the reasons that lead to algorithmic discrimination.
We also discuss 
fairness mitigation approaches categorized according to three stages of deep learning life-cycle,
aiming to push forward the area of fairness in deep learning and build genuinely fair and reliable deep learning systems.
\end{abstract}

}

\maketitle

\IEEEdisplaynontitleabstractindextext

%
\IEEEpeerreviewmaketitle

\IEEEraisesectionheading{\section{Introduction}\label{sec:introduction}}

%
%
%
%
\IEEEPARstart{M}{achine} 
learning algorithms have achieved dramatic progress nowadays,
and are 
increasingly being deployed in high-stake applications, 
including employment, 
criminal justice, personalized medicine, \emph{etc}~\cite{gajane2017formalizing}. Nevertheless, \emph{fairness in machine learning} remains a  problem.
Machine learning algorithms have the risk of amplifying societal stereotypes by over associating protected attributes, e.g., race and gender, with the prediction task~\cite{wang2018adversarial}.
Eventually they are capable of exhibiting discriminatory behaviors against certain subgroups. For example, a recruiting tool for STEM jobs believes that men are more qualified and shows bias against women~\cite{kiritchenko2018examining},
facial recognition performs extremely poorly for female with darker skin~\cite{buolamwini2018gender}, 
recognition accuracy is very low for subgroup of people in pedestrian detection of self-driving cars~\cite{wang2018adversarial}. 
The fairness problem might cause adverse impacts on individuals and society. It not only limits a person's opportunity that s/he is qualified, but also might further exacerbates social inequity. 

Among different machine learning models, the fairness problem of \emph{deep learning models} has  attracted attention from academia and industry recently. First, deep learning models have 
achieved the state-of-the-art performance in many domains. Their success can partially be attributed to the data-driven learning paradigm, which enables the models to learn useful representations automatically from data. The data might contain human biases, which reflect historical prejudices against certain social groups and existing demographic inequalities. 
The data-driven learning also inevitably causes deep learning models to replicate and even amplify biases present in data. 
Second, it remains a challenge to diagnose and address the deep learning fairness problem. 
Deep learning models are generally regarded as black-boxes, and their intermediate representations are opaque and hard to comprehend. This is problematic and makes it
difficult to identify whether these models make decisions based on right and justified reasons, or  due to biases. In addition, this makes it challenging to design bias detection and mitigation approaches.



In this article, we summarize \emph{fairness in deep learning} work from the computational perspective, and do not discuss work from social science, law and many other disciplines~\cite{gajane2017formalizing}. 
Particularly we show that interpretability could significantly contribute to better understandings of the reasons that affect fairness. We also review fairness mitigation strategies categorized into three stages of deep learning life-cycle.
Finally, we propose open challenges and future research directions. 
Throughout this article, \emph{we don't differentiate between deep learning and DNN} (Deep neural network) and use them interchangeably. 
Besides, we abstract from the exact  
DNN architectures, e.g., convolutional neural network (CNN), 
recurrent neural network (RNN), and multi-layer perceptron (MLP), 
and focus more on conceptual aspects which underlie the success of DNN bias detection and mitigation techniques.

\begin{table*}
\caption{DNN fairness problem categorization and representative examples.}
\small
\centering
\scalebox{0.96}{
\begin{tabular}{l l}
\toprule
\textbf{Class} &\textbf{Representative examples}\\
\midrule
\textbf{Discrimination via Input }& \emph{Employment}: Recruiting tool believes that men are more qualified and shows bias\\
& \,  against women. \\
 & \emph{Loan Approval}: Loan eligibility system negatively rates people belonging to certain  \\ 
 & \, ZIP code, causing discrimination for certain races. \\
 & \emph{Criminal Justice}: Recidivism prediction system predicts black inmates are three times  \\
 \vspace{1.5mm}
 &\, more likely to be classified as `high risk' than white inmates.\\

\textbf{Discrimination via Representation}& \emph{Medical Image Diagnosis}: CNN model could identify patients' self-reported sex from a \\  
& \, retina image, and shows discrimination based on gender.\\
& \emph{Credit Scoring}:  Using raw texts as input, demographic information of authors is encoded
\\
\vspace{1.5mm}
& \, in the intermediate representations DNN-based credit scoring classifiers. \\

\textbf{Prediction Quality Disparity}& \emph{Facial Recognition}: Facial recognition performs very poorly for female with darker skin.\\
& \emph{Language processing}: Language identification models perform significantly  worse when \\
& \,   processing text produced by people belonging to certain races.\\
& \emph{Readmission}: ICU mortality and psychiatric 30-day readmission model prediction accuracy \\
& \,  is significantly different across gender and insurance types. \\
\bottomrule
\end{tabular}
}
\label{tab:example}
\end{table*}

\section{DNN Fairness}
In this section, we introduce the 
categorization of fairness problem, measurements of fairness, and interpretation methods closely relevant to understanding DNN fairness.

\subsection{Fairness Problem Categorization}
From the computational perspective, DNN unfairness can be generally categorized into two classes: 
\emph{prediction outcome discrimination}, and \emph{prediction quality disparity}.

\subsubsection{Prediction Outcome Discrimination}

\begin{figure*}
  \centering
  \includegraphics[width=1\linewidth]{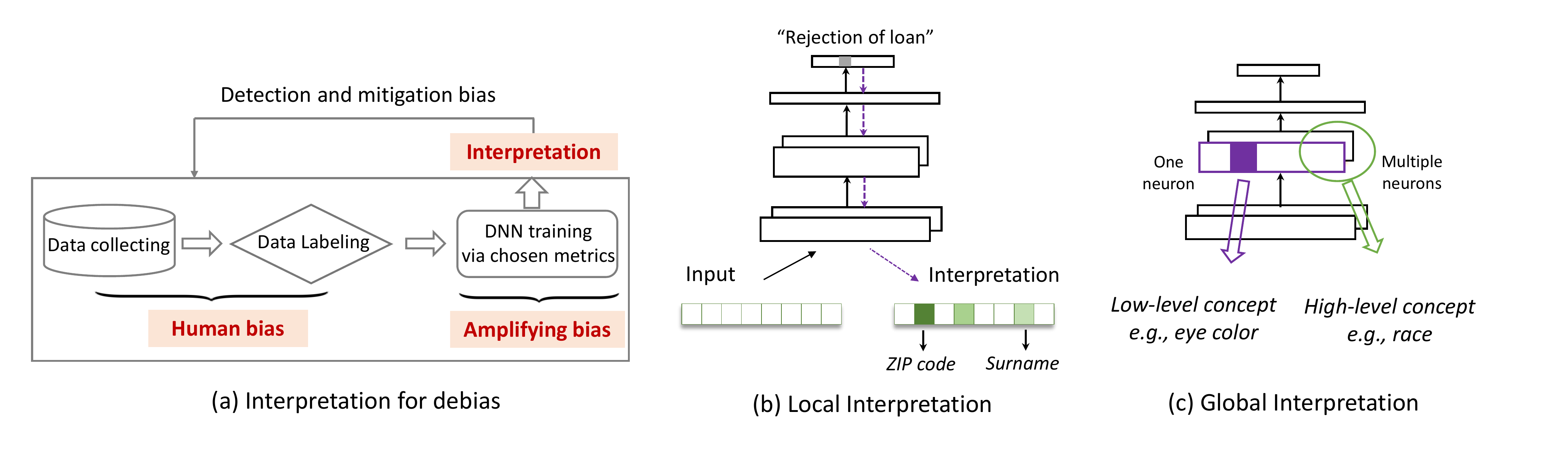}
  \vspace{-8mm}
  \caption{(a) Bias exists in different stages of the DNN training pipeline, and interpretation could be utilized to detect and mitigate bias. (b) DNN local interpretation, (c) DNN global interpretation.}
  \label{fig:overall}
\end{figure*}

Discrimination refers to the phenomenon that DNN models produce unfavourable treatment of people due to the membership of certain demographic groups~\cite{gajane2017formalizing}. For instance, a recruiting tool believes that men are more qualified and shows bias against women, and loan eligibility system negatively rates African Americans.
Current DNNs generally follow the purely data-driven and end-to-end 
learning paradigm, which are trained with labeled data. 
The model training pipeline is illustrated in Fig.~\ref{fig:overall}(a).
Any training data may contain some biases, either intrinsic noise or additional signals inadvertently introduced by human annotators~\cite{gururangan2018annotation}.  
DNNs are designed to fit these skewed training data, and thus would naturally replicate the biases  existed in data.
 Even worse, 
DNNs not only rely on these biases to make decisions, but also 
make unwanted implicit associations and amplify societal stereotypes about people~\cite{bolukbasi2016man,zhao2017men}. 
This eventually results in trained models with algorithmic discrimination.
Outcome discrimination can be further split into \emph{Input} and \emph{Representation} prospective. 
We present below detailed descriptions for these 
two subcategories, with representative examples in Tab.~\ref{tab:example}.

\vspace{1mm}
\noindent\textbf{Discrimination via Input} \,
Prediction outcome discrimination could be traced back to the input.
Even though a DNN model does not explicitly take \emph{protected attributes} as input, e.g., race, gender and age, 
it may still induce 
prediction discrimination~\cite{kallus2019assessing}. In the context of DNN systems, \emph{protected attributes are often not observed in the input data}, mainly due to two reasons. Firstly, most DNN models rely on raw data, e.g., text, 
as input and thus protected attributes are not explicitly encoded in the input. Secondly, collecting protected attributes 
such as race and ethnicity information 
is often not allowed by the law in real-world applications. Despite the absence of explicit protected attributes, DNNs still could exhibit unintentional discrimination,
since there are some features highly correlated with class membership. For instance, ZIP code and surname could indicate race, many words within text input could be used to infer gender~\cite{kallus2019assessing}. 
The model prediction might highly depends on the class memberships, and eventually shows discrimination to certain demographic group.

\vspace{1mm}
\noindent\textbf{Discrimination via Representation} \,
Sometimes prediction outcome discrimination needs to be diagnosed and mitigated from the representation prospective.
In some cases, attributing the bias to input is nearly impossible, e.g., for image input. For instance, CNN model could identify patients' self-reported sex from a retina image, while humans even ophthalmologists cannot identify cues from the input image. 
Besides, in some scenarios, finding the sensitive input attributes are challenging if the input dimension is too large~\cite{beutel2017data}. 
In those settings, different demographic groups would have distinct DNN intermediate representations. The class memberships of different protected attributes 
could be encoded in deep representations.  DNN model will make decisions based on the implicitly learned membership information and produce discriminate classification outcomes. Thus prediction outcome discrimination could be detected and removed from the deep representation perspective.

\subsubsection{Prediction Quality Disparity}
Prediction quality difference of models for different protected groups is another important category of unfairness.
DNN systems have shown lower quality for some groups of people as opposed to other groups. 
Different from prediction outcome discrimination which is about \emph{resources and opportunities allocations harm} in high-stake applications such as hiring, loan and credit, this category is about \emph{quality of services harm} that usually happen in general applications, e.g., facial recognition and language processing (See Tab.~\ref{tab:example}).
Examples include 
the language identification systems perform significantly worse when processing text produced by people belonging to certain races~\cite{blodgett2016demographic,jurgens2017incorporating}, health care applications including ICU mortality and psychiatric 30-day readmission model prediction accuracy is significantly different across gender and insurance types~\cite{chen2019can}. 
This is usually due to the underrepresentation problem,
where data may be less informative or less reliably collected for certain parts of the population. 
Take the Imagenet dataset (ILSVRC 2012 subset with 1000 categories) 
as an example: females comprise only 41.62\% of images, people over 60 are almost non-existent~\cite{dulhanty2019auditing}.
The typical objective of DNN training is to minimize the overall error. If the model cannot simultaneously fit all populations optimally, it will fit the majority group. Although this may maximize overall prediction accuracy, it comes at the expense of the under-represented populations and leads to their poor performance.

\subsection{Measurements of Fairness}\label{measurements-of-fairness}
Many different metrics have been proposed to measure the fairness of machine learning models. 
One line of work measures \emph{individual fairness}, which follows the philosophy that similar inputs should yield similar predictions~\cite{dwork2012fairness}. Nevertheless, this leaves the open question of how to define input similarity~\cite{beutel2019putting,gajane2017formalizing}. Another line of work focus on \emph{group fairness}, where examples are grouped according to a particular sensitive attribute, and statistics about model predictions is calculated for each group and compared across groups~\cite{beutel2019putting}. Comparing to individual fairness, group fairness is more widely adopted in fairness research, and thus is the focus of this article.
Different kinds of group fairness measurements have been proposed, and 
we will introduce below three mostly used ones.

\vspace{1mm}
\noindent\textbf{Demographic Parity} \, 
It asserts that average of algorithmic decisions should be similar across different groups: 
$\frac{p(\hat{y}=1|z=0)}{p(\hat{y}=1|z=1)}\geq \tau $, where $\tau$ is a given threshold, usually set as 0.8~\cite{feldman2015certifying},
$\hat{y}$ is a model prediction, 1 denotes favorable outcome, $z$ denotes \emph{protected attribute}, e.g., race, gender.
Demographic parity is independent of the ground truth labels. This is useful especially when reliable ground truth information is not available, e.g., employment, credit, and criminal justice~\cite{gajane2017formalizing}.

\vspace{1mm}
\noindent\textbf{Equality of Opportunity} \,
This metric has taken into consideration that different groups could have different distribution in terms of label $y$. It is defined as: ${p(\hat{y}=1|z=0, y=1)}-{p(\hat{y}=1|z=1, y=1)}$, where $y$ is the ground truth label~\cite{hardt2016equality}. Essentially this is comparing the \emph{true positive} rate across different groups. A symmetric measurement can be calculated for \emph{false positive} rate: ${p(\hat{y}=1|z=0, y=0)}-{p(\hat{y}=1|z=1, y=0)}$. Putting them together will result the \emph{Equality of Odds} metric~\cite{hardt2016equality}.

\vspace{1mm}
\noindent\textbf{Predictive Quality Parity} \,
This metric measures prediction quality difference between different subgroups. 
The quality denotes quantitative model performance in terms of model predictions and ground truth, 
and in this work we focus on \emph{accuracy} measurement for multi-class classification~\cite{buolamwini2018gender}. 
It is desirable that a model has equal prediction accuracy across different demographic subgroups.

For a more comprehensive discussion of measurements, we refer interested readers to the work~\cite{gajane2017formalizing}. It is worth noting that different applications require different measurements which satisfy their specific ethical and legal requirements.



\subsection{Interpretability for Addressing Fairness Problem}
DNNs are often regarded as black-boxes and criticized by the lack of interpretability, since these models cannot provide meaningful interpretation on how a certain prediction is made.  
Interpretability could be utilized as an effective debugging tool to 
analyze the models, and enhance the transparency and fairness of models (Fig.~\ref{fig:overall}(a)).
Interpretability can generally be grouped into two categories: local interpretation and global interpretation~\cite{du2018techniques}. 

\vspace{1mm}
\noindent\textbf{Local Interpretation}\,
Local interpretation could illustrate how the model arrives at a certain prediction for a specific input (Fig.~\ref{fig:overall}(b)).
It is achieved by attributing model's prediction in terms of its input features. The final interpretation is illustrated in the format of \emph{feature importance} visualization~\cite{du2018towards,du2019on}. Take loan prediction for example. The model input is a vector containing categorical features, and the interpretation result is a heat map(or attribution map), 
where features with higher scores represent higher relevance for the prediction.

\vspace{1mm}
\noindent\textbf{Global Interpretation}\,
The goal is to provide a global understanding about what knowledge has been captured by a pre-trained DNN, and illuminate the learned representations in an intuitive manner to humans (Fig.~\ref{fig:overall}(c)). 
The simplest way is to comprehend the concept captured by a single neuron,
which is the representation derived from a specific channel at a specific layer~\cite{yosinski2015understanding}. 
The combination of multiple neurons of different channels or even different layers could represent more abstract concepts~\cite{kim2017interpretability}. 
Those protected concepts are usually based on multiple elementary low-level concepts. For instance, race concept can be indicated via multiple local clues such as eye color and hair color. 
Thus comparing to concepts learned by a single neuron, concepts yielded by a combination of neurons are more relevant to fairness.


\section{Detection of Modeling Bias} 
In this section, we present methods for detecting and understanding algorithmic discrimination,   
by making use of DNN interpretability as an effective computational tool.

\begin{figure*}
  \centering
  \includegraphics[width=0.9\linewidth]{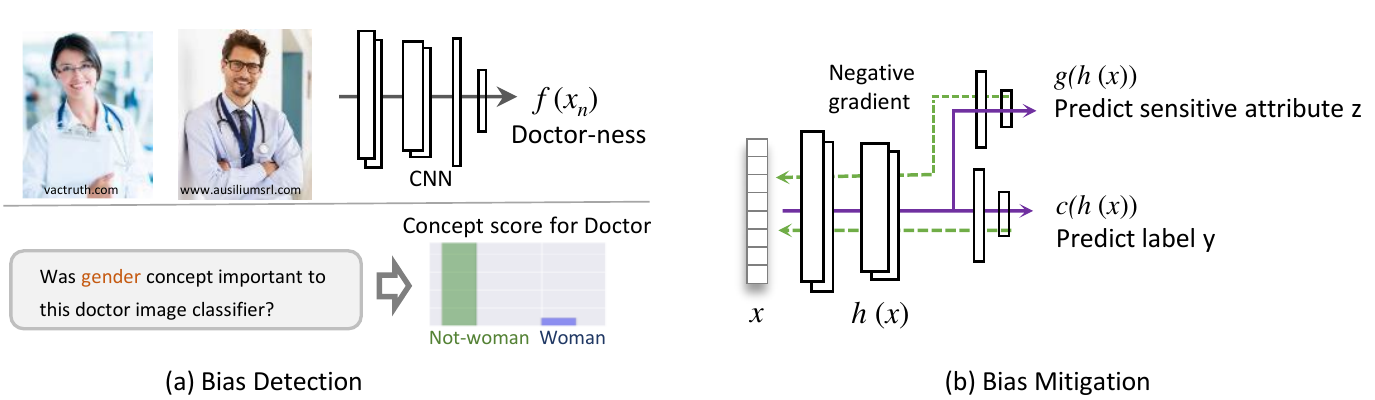}
  \vspace{-2mm}
  \caption{(a) Global interpretation for detection of discrimination. Results show that this CNN has captured gender concept, and the \emph{not-woman} concept would significantly increase doctor prediction confidence of the CNN classifier. Thus it indicates the CNN's discrimination towards woman. (b) Adversarial training for mitigation of discrimination. The intuition is to enforce deep representation to maximally predict main task labels, while at the same time minimally predict sensitive attributes.}
  \label{fig:overall2}
\end{figure*}

\subsection{Discrimination via Input} \label{inputdetection}
The source of prediction outcome discrimination 
could be traced back to the input features.
As discussed in Sec.~2.1.1, protected attributes
are often not observed in the input data. Due to the redundant encodings, other seemingly innocuous features may be highly correlated with protected attribute and 
cause model bias~\cite{hardt2016equality}. 
The goal here is to locate these features via local DNN interpretation.

The first solution is performed in a top-down manner, where local interpretation is employed to generate feature importance vector.
After getting feature importance for all input features, we can take out those with relatively high importance scores and further analyze them.
Among this subset of features, the focus is to identify those fairness sensitive features (in contrast to task relevant features). Take the loan application for example. If the features contributing most to DNN prediction include surname and ZIP code of applicants, we can assert that this model has discrimination towards certain race, and surname and ZIP code here are fairness sensitive features (Fig.~\ref{fig:overall}(b)). 
The second solution is implemented in the bottom-up manner. Humans first pre-choose features which they are skeptical to be associated with protected attributes, and then analyze feature importance of the identified features~\cite{kiritchenko2018examining}.
These subset of features are perturbed to generate new data samples, i.e., counterfactual(s)~\cite{sharma2020certifai}.
We then feed the counterfactual to the DNN and observe the model prediction difference. If the perturbation of those suspected fairness sensitive features causes significant model prediction change, it can be asserted that the DNN has made biased decisions based on protected attributes. 
Note that statistical differences are calculated over a set of similar instances, so as to validate whether the model has violated \emph{group fairness}.

A representative example is using local interpretation to detect race bias in sentiment analysis systems~\cite{kiritchenko2018examining}. 
Common African American first names (e.g., Malik)
and European American first names (e.g., Ellen)
are chosen as sensitive features. The comparison is between average prediction scores of sentences with first name of these two races. 
The results show statistically significant race bias, where DNNs consistently yield higher sentiment prediction with African American name on the tasks of anger, fear and sadness intensity prediction. 
The results indicate the models have violated \emph{demographic parity} metric, and reflect the stereotypes that African Americans are relevant to negative emotions.

\subsection{Discrimination via Representation} 
Sometimes it is hard to identify bias from the input perspective, and detecting model bias from the deep representations is more convenient.
DNN global interpretation could be exploited as a debugging tool to analyze the deep representations. 
The goal is to identify whether a protected attribute has been captured by the intermediate representation, and the degree to which this protected attribute contributes to the model prediction. Thus a two-stage scheme could be applied to detect discrimination.


Firstly, global interpretation is utilized to analyze whether a DNN has learned a protected concept. This is usually achieved by pointing to a direction in the activation space of DNN's intermediate layers~\cite{kim2017interpretability,zhou2018interpretable,fong2018net2vec}. A typical example 
is the \emph{concept activation vector (CAV)} method~\cite{kim2017interpretability}.
Here CAV defines a high-level concept using a set of example inputs. For example, to define concept \emph{African American race}, a set of darker skin Congoid images could be used. The CAV vector is the direction of activation values for the set of examples corresponding to that concept. This vector is obtained by training a linear classifier between the concept examples and a set of random counterexamples, where the vector is the direction orthogonal to the decision boundary. 
Secondly, after confirming that a DNN has learned a protected concept, we proceed to test the contribution of this concept towards model's final prediction. Different strategies can be used to quantify the conceptual sensitivity, including the top-down manner which calculates derivative of DNN's prediction to the concept vector~\cite{kim2017interpretability}, or the bottom-up manner which adds this concept vector to different inputs' intermediate activation and then observe the change of model predictions. Ultimately the representation bias level for a protected attribute is described using a numerical score. The higher of the numerical sensitivity score, the more significantly that this concept contributes to DNN's prediction. 

A representative example is detecting the gender bias in deep representations of a CNN doctorness classifier (Fig.~\ref{fig:overall2}(a)).
TCAV indicates that the model indeed has captured the gender concept. Besides, \emph{not-woman} concept would dramatically increase the model's prediction confidence of doctorness. The findings have conformed the CNN's discrimination towards women. They also reflect the commonly held gender stereotype that doctors are men. 

\subsection{Prediction Quality Disparity} 
There usually happens that some groups appear more frequently than others in training data. The DNN model will optimize for those groups in order to boost the overall model performance, leading to  
low prediction accuracy for the minority group.

The detection of prediction quality disparity is typically performed in a two-step manner: splitting data into subgroups according to sensitive attributes, and calculating the accuracy for each demographic groups.
For instance, facial recognition systems are analyzed in terms of their prediction quality~\cite{buolamwini2018gender}. Human face images are classified into four categories: darker skin males, darker skin females, lighter skin males, and lighter skin females. Three gender classification systems are evaluated for the four groups, and substantial accuracy disparities are observed. For all three systems, the darker skin females group yields the highest mis-classification rate, with error rate up to 34.7\%. In contrast, the maximum error rate for lighter skin males is 0.8\%. These results conform that the model has violated \emph{predictive quality parity} metric and raise an urgent need for building fair facial analysis systems.

Beyond the verification accuracy, model interpretability could be used to analyze the reasons of discrimination. A decomposition-based local DNN interpretation method, i.e., \emph{class activation maps (CAM)}~\cite{zhou2016learning}, is used to 
investigate the regions of interest attended by the DNN models when making decisions. CAM is utilized to analyze two groups: lighter skin and darker skin group~\cite{nagpal2019deep}. The visualization shows that the model needs to focus on eye region for lighter skin group, while focus on the nose region 
and chin region for darker skin group. It suggests different strategies are needed to make decisions for different demographic groups. If the training dataset has inadequate samples for darker skin group, the trained model may capture representation preference for the majority group and fail to learn effective classification strategy for minority darker skin group, thus leading to poor performance for minority group.

\begin{table*}
\caption{Representative algorithms for mitigating unfairness in DNN models. Pre-processing, in-processing, and post-processing correspond to three stages of deep learning pipeline: dataset construction, model training, and model inference.}
\vspace{-2mm}
\small
\centering
\scalebox{0.9}{
\begin{tabular}{l l l l}
\toprule
\textbf{Class} &\textbf{Pre-processing } &\textbf{In-processing } &\textbf{Post-processing }\\
\midrule
\textbf{Discrimination  }& Sensitive features removal & Attribution regularization~\cite{ross2017right,liu2019incorporating} & Calibrated distribution~\cite{zhao2017men} \\
\textbf{via Input} & Sensitive features replacement & Reduction game~\cite{agarwal2018reductions} & Calibrated equalized odds~\cite{hardt2016equality} \\
& Reweighing~\cite{kamiran2012data}  & Prejudice remover~\cite{kamishima2012fairness} &  \\
\vspace{1mm}
 & Optimized pre-processing~\cite{calmon2017optimized} &  &  \\
\midrule 

\textbf{Discrimination} & Balanced dataset collection & Adversarial training~\cite{wang2018adversarial,elazar2018adversarial} & Troubling neurons turn off \\
\textbf{via Representation} &  & Adversarial fairness desideratum~\cite{madras2018learning} &  \\
&  & Semantic constraints~\cite{quadrianto2019discovering} &  \\
\vspace{1mm}
&  & Distance metrics~\cite{louizos2015variational,jiang2019wasserstein} &  \\
\midrule

\textbf{Prediction Quality}& Diverse dataset collection~\cite{escalera2016chalearn} & Transfer learning~\cite{ryu2017inclusivefacenet} &  \\
\textbf{Disparity}& Synthetic data generation~\cite{zhang2017age} & Multi-task learning~\cite{das2018mitigating} &  \\
\bottomrule
\end{tabular}
}
\label{tab:mitigation}
\end{table*}

\section{Mitigation of Modeling Bias}
After presenting bias detection approaches, 
we introduce below methods which could mitigate against 
adverse biases.
A typical and simplified deep learning pipeline could be split into three stages: dataset construction, model training, and inference. Mitigation methods could be correspondingly divided into three broad groups: pre-processing, in-processing, and post-processing~\cite{bellamy2018ai} (see Tab.~\ref{tab:mitigation}).
Pre-processing tries to debias and increase the quality of training set. In-processing adds auxiliary regularization term to the overall objective function during training, explicitly or implicitly enforcing constraints for certain fairness metric. Post-processing is performed after model training to calibrate the predictions of trained models.

\subsection{Discrimination via Input}
In this section, we introduce some representative mitigation methods as well as their empirical evaluation.

\subsubsection{Pre-processing}
A straightforward solution is to remove those fairness sensitive features from training data. 
For instance, surname and ZIP code can be deleted to reduce the 
discrimination of DNNs towards certain race.
A drawback of directly removing features is that this might lead to poor model performance and thus reduces model utility. 
We can replace these fairness sensitive features with alternative values. Take the sentence `\emph{The conversation with Malik was heartbreaking}' for example, we can replace \emph{`Malik'} with \emph{`IDENTITY'} to reduce the possibility that DNN model shows discrimination based on races.
In addition, there are some data-agnostic pre-processing transformation techniques. For instance, the weights of each training sample is given differently to ensure fairness before model training~\cite{kamiran2012data}. Another transformation is formulated in a probabilistic
framework, where features and labels are edited to ensure group fairness~\cite{calmon2017optimized}. 

\begin{figure}
  \centering
  \includegraphics[width=0.98\linewidth]{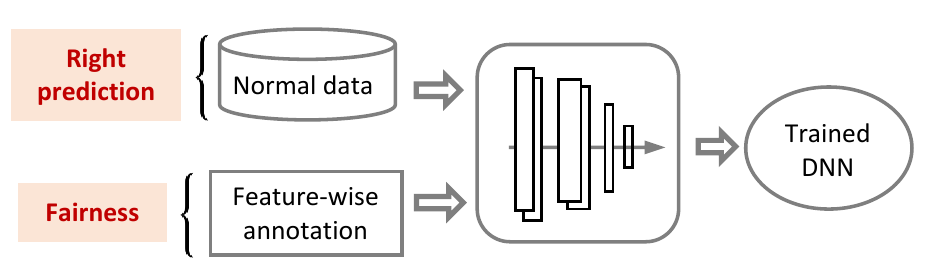}
  \caption{Using regularization for mitigation of discrimination via input. Besides normal training data and ground truth, feature-wise annotations are also needed, specifying which subset of features is fairness sensitive, and which subset is task relevant. The key idea is to enforce DNNs to depend less on sensitive features. }
  \label{fig:rationale-training}
\end{figure}

\subsubsection{In-processing}
An alternative approach to mitigate discrimination is via model regularization. The regularization implicitly or explicitly optimizes a fairness metric.

\vspace{2mm}
\noindent\textbf{Implicit Regularization}: The first category adds implicit constraints which disentangle the association between model prediction and fairness sensitive attributes (Fig.~\ref{fig:rationale-training}). It
enforces DNN models to pay more attention to correct features relevant to prediction task, rather than capture spurious correlations between prediction task and protected attributes. 
Specifically, the model training is regularized with local DNN interpretation~\cite{ross2017right,liu2019incorporating,du2019learning}. Beyond ground truth $y$ for the input $x$, the regularization also needs feature-wise annotations $r$, specifying whether each feature within the input correlates with protected attributes or not. Annotation $r$ either could be labelled by domain experts or identified through the detection methods in Sec.~\ref{inputdetection}. For instance, the annotation $r$ for input `\emph{The conversation with Malik was heartbreaking}' is $[0, 0, 0, 1, 0, 0]$, indicating that `Malik' is correlated with race, while the rest words are considered as task relevant. 
The overall loss function is denoted as:
\begin{equation}
L(\theta, x, y, r) = \underbrace{d_1(y, \hat{y})}_{Prediction} +  \underbrace{\lambda_1  d_2(f_{loc}(x),r)}_{Fairness}+ \underbrace{\lambda_2 \mathcal{R}(\theta)}_{\small{Regularizer}}, 
\end{equation}
where $d_1$ is normal classification loss function, e.g., cross entropy loss, and $\mathcal{R}(\theta)$ is a regularization term. Function $f_{loc}(x)$ is local interpretation method, and $d_2$ is a distance metric function. The three terms are used to guide the DNN model to make right prediction, make decision based on right and unbiased evidences, and not overfit to training set respectively. Hyperparameters $\lambda_1$ and $\lambda_2$ are used to balance three terms. Note that 
$f_{loc}(x)$ needs to be end-to-end differentiable, amenable for training with back-propagation and updating DNN parameters. 
The resulting fair model depends more on holistic information which is task relevant, while at the same time
conditions less on sensitive attributes. Besides, the trained models also satisfy better \emph{demographic parity} criteria.

\vspace{1mm}
\noindent\textbf{Explicit Regularization}: This category adds explicit constraints through updating model's loss function to minimize the performance difference between different groups~\cite{kamishima2012fairness,agarwal2018reductions}. They optimize the trade-off between accuracy and a specific kind of fairness metric given training-time access to protected attributes. A representative example combines demographic parity and equality of odds into overall objective function~\cite{agarwal2018reductions}. Specifically, it define a “reduction” that treats the accuracy-fairness trade-off as a sequential game between two players. At each step in the gaming sequence, one player maximizes accuracy and the other player imposes a particular amount of fairness. This method is advantageous in that it is model agnostic and could be applied to different DNN architectures.

\subsubsection{Post-processing}
Post-processing calibration takes the model's prediction and protected attribute to calibrate model's prediction during the inference time~\cite{zhao2017men,hardt2016equality}. The goal is to enforce prediction distribution to approach either the training distribution, or a specific fairness metric. Firstly, corpus-level constraints are utilized to enforce the model prediction distribution to follow the training data distribution~\cite{zhao2017men}.
Secondly, the calibration could also be performed towards a fairness metric. For instance, one technique takes as input an existing classifier and the sensitive feature, and derives a monotone transformation of the classifier's prediction to enforce the specified equalized of odds constraint~\cite{hardt2016equality}. These methods allow for diverse fairness metrics and prove to be effective to reduce discrimination. On the other hand, these methods could be problematic since they require inference-time access to protected attributes, which however usually are not available during inference time in real-world applications.

\subsubsection{Evaluation of Mitigation Algorithms}
We conduct experiments to evaluate performance of different mitigation algorithms. We use \emph{Adult Census Income}~\footnote{\url{https://archive.ics.uci.edu/ml/datasets/Adult}} and \emph{COMPAS}~\footnote{\url{https://github.com/propublica/compas-analysis}} two datasets, containing 48,842 and 6,167 instances respectively. We use gender and race as protected attribute for the two datasets respectively. Each dataset is split into 50\% for training, 20\% for validation and 30\% for testing.  The base DNN model is a multilayer perceptron (MLP) with 3 layers~\footnote{\url{https://scikit-learn.org/stable/modules/generated/sklearn.neural_network.MLPClassifier.html}}. We evaluate the metrics with the best performing model on validation set. The results are displayed in Tab.~\ref{tab:fairnessEvaluation}. Note that we average each number over three runs, to eliminate influence of random initialization to DNN performance. We have the following key observations. Firtly, without using the debiasing algorithms, DNN models would amplify bias existing in training data, as shown by the comparison of Parity value between DNN and training set. Secondly, there is fairness utility trade-off, where most mitigation algorithms could compromise overall model accuracy. Thirdly, fairness measurements could be conflicting with others. Some mitigation methods may be fair in terms of demographic parity, but may result in unfairness with regard to equality of opportunity/odds. Fourthly, mitigation could possibly lead to discrimination towards majority groups, where equality of opportunity/odds metrics switch from negative values to positive values. 


\begin{table}
\centering
\setlength{\tabcolsep}{4pt}
\caption{Mitigation comparison between 5 methods for discrimination via input. For accuracy and demographic parity, the close to 1 the better. For equality of opportunity and equality of odds, the close to 0 the better.}
\scalebox{0.82}{
\begin{tabular}{l c c c c c c c c}
\toprule 
& \multicolumn{4}{c}{\textbf{Adult Income}} & \multicolumn{4}{c}{\textbf{COMPAS}} \\
\cmidrule(l){2-5} \cmidrule(l){6-9}
\textbf{Model/Data} & Acc & Parity & Opty & Odds  & Acc & Parity & Opty & Odds\\ 
\midrule 
Dataset\_bias & n/a & 0.386 & n/a & n/a  & n/a & 0.747 & n/a & n/a\\ 
DNN\_original & 0.836 & 0.347 & -0.094  & -0.089 & 0.658 & 0.741 & -0.160 &-0.136\\ 
\midrule 
Reweighting~\cite{kamiran2012data} & 0.832 & 0.654 & -0.106  & -0.090 & 0.652 & 0.788 & -0.186 & -0.149\\ 
Optimized\_pre~\cite{calmon2017optimized} & 0.778 & 0.573 & -0.107 & -0.088 & 0.665 & 0.959 & -0.018 & -0.024\\ 
Prejudice\_rem~\cite{kamishima2012fairness} & 0.817 & 0.961 & 0.005  & 0.039 & 0.635 & 0.937 & 0.008 & 0.062\\ 
\footnotesize{Calibrated\_odds}~\cite{hardt2016equality} & 0.804 & 0.546 & 0.148  & 0.052 & 0.639 & 0.819 & 0.036 & 0.150\\
\bottomrule 
\end{tabular}
}
\label{tab:fairnessEvaluation}
\end{table}

\subsection{Discrimination via Representation} 
The goal is to reduce representation bias while at the same time preserve useful prediction properties of DNNs. 

\subsubsection{Pre-processing}
Collecting balanced dataset is a possible way is alleviate representation bias, since 
prediction discrimination is partially caused by difference of label distribution conditioning on protected features in the training data. Take text dataset for example, gender swapping can be used to create a dataset which is identical to the original one but biased towards another gender. The union of the original and gender-swapping dataset would be gender balanced, which can be used to retrain DNN models. 
However, it is still not guaranteed that balanced dataset could eliminate the representation bias.
Previous studies show that even training data is balanced,
DNNs still could capture information like gender, race in intermediate representation~\cite{elazar2018adversarial,wang2018adversarial}. Thus more fundamental changes in DNN models are needed to further reduce discrimination.

\subsubsection{In-processing}
\textbf{Adversarial Learning:}
From model training perspective, adversarial training~\cite{ganin2016domain} is a representative solution to remove information about sensitive attributes from intermediate representation of DNNs~\cite{elazar2018adversarial,wang2018adversarial}. 
A predictor and an adversarial classifier are learned simultaneously. The goal of the predictor is to learn a high-level representation which is maximally informative for the major prediction task, while the role of adversarial classifier is to minimize the predictor's ability to predict the protected attribute (Fig.~\ref{fig:overall2}(b)). 
The DNN is denoted as $f(x) = c(h(x))$, where $h(x)$ is the intermediate representation for input $x$, and $c(\cdot)$ is responsible to map intermediate representation to final model prediction. 
The protected attribute 
is denoted using $z$.
An adversarial classifier $g(h(x))$ is also constructed to predict protected attribute $z$ from representation $h(x)$. The adversarial training process is denoted as follows:
\begin{equation}
\begin{aligned}
\centering
& \text{arg} \, \underset{ g }{\text{min}} \quad  L(g(h(x)),z) \\
& \text{arg} \, \underset{ h,c }{\text{min}} \quad  L(c(h(x)),y) - \lambda L(g(h(x)),z),
\end{aligned}
\end{equation}
where the adversarial classifier is to penalize the representation of $h(x)$ if protected attribute $z$ is predictable, parameter $\lambda$ is used to negotiate the trade-off between maximizing utility and fairness. The training is iteratively performed between the main classifier $f(x)$ and the adversarial classifier $g(h(x))$. 
Some methods implement adversarial using general cross entropy loss~\cite{wang2018adversarial,elazar2018adversarial}, while some others use advanced adversarial objectives according to fairness desideratum~\cite{madras2018learning}.
The adversarial frameworks show improved performance on metrics like \emph{demographic parity} and \emph{equality of opportunity}. In the meantime, adversarial training has some pitfalls. Firstly, it could not fully retain the semantic meaning of the data~\cite{quadrianto2019discovering}, thus could harm model \emph{accuracy}, especially when adding a strong regularization, i.e., a large $\lambda$. Secondly, it is also hard to stabilize the training, similar like adversarial training in other applications. 


\vspace{2mm}
\noindent\textbf{Beyond Adversarial Learning:} Besides adversarial framework, some other advanced fair representation learning methods are proposed recently. For instance, residual decomposition is used for fair representation learning~\cite{quadrianto2019discovering}. Beyond enforcing inner representations to suppress protected attribute and predict the main task label, this method also adds regularization term to ensure that the debaised representation lies in the same space with original input. With the semantic meaning constraints, such representation learning methods thus could achieve better trade-off between fairness metrics and accuracy. In addition, there exist non-adversarial methods using distance metrics, such as maximum mean discrepancy~\cite{louizos2015variational} and Wasserstein distance~\cite{jiang2019wasserstein}, aiming to learn fair representations and eliminate disparities between different sensitive groups.

\subsubsection{Post-processing}
The mitigation of discrimination through representation could also be implemented at the inference stage. The key idea is to suppress the neurons that have captured protected attributes. The process is split into two-stages. Firstly, global interpretation methods are used to locate the neurons that are highly related with protected attributes~\cite{kim2017interpretability}. Secondly, the activation values flowing out from those neurons are set to zero, so as to turn off the correlation between protected attribute and DNN model prediction.

\subsection{Prediction Quality Disparity} 
In this section, we introduce methods which could increase the prediction quality for underrepresented minorities.

\subsubsection{Pre-processing}
From data perspective,
one straightforward idea to increase the prediction quality of underrepresented group is to enforce the training dataset to be diverse.
This can be achieved by collecting data from more comprehensive data sources. For instance, the \emph{Faces of the World} dataset is developed, aiming to achieve a uniform distribution of face images across two genders and four ethnic groups~\cite{escalera2016chalearn}.
In some domains collecting data might be expensive or impractical,
and Generative Adversarial Networks (GANs) could be used to generate synthetic data~\cite{frid2018synthetic}. 
For instance, GANs are utilized to generate face images across all age ranges~\cite{zhang2017age}. DNN models trained on this dataset are able to achieve equal predictive quality even for those previously minority age groups, e.g., age over 60.

\subsubsection{In-processing}
Regularizing model training is another perspective to increase accuracy of the minority groups.
This could be implemented using the transfer learning framework. For instance, transfer learning is proposed to solve the problem of unequal face attribute detection performance across different race and gender subgroups~\cite{ryu2017inclusivefacenet}. A CNN model is firstly trained using source domain dataset which is rich with data for the minority group, i.e., the aforementioned \emph{Faces of the World} dataset. Then the trained CNN model is transferred to the target domain, i.e., 
face attribute detection, to improve accuracy of the minority group. 
Transfer learning could promote both the overall accuracy and gender/race subgroup accuracy. Model training regularization could also be achieved under the multi-task learning setting. For instance, a multi-task learning framework is designed for joint classification of gender, race, and age of faces images~\cite{das2018mitigating}. This can yield significant accuracy improvement for different demographic subgroups, thus promoting model fairness in terms of \emph{predictive quality parity} measurement.

\section{Research Challenges}
Despite significant progresses for fairness in deep learning, there are still some challenges deserving further research.

\vspace{2mm}
\noindent\textbf{Benchmark Datasets}\, 
Benchmark datasets are lacking to systematically examine the inappropriate biases in trained DNN systems~\cite{kiritchenko2018examining}. Benchmark dataset means the dataset which has been teased out biases towards certain protected groups. Current practice of testing DNNs' performance is using hold-out test sets, which usually contain the same biases as training set.
Test sets might fail to unveil the unfairness problem of trained models. Each DNN is recommended to evaluate its fairness on benchmark datasets, serving as supplementary test sets beyond normal test set. To facilitate the construction of benchmark datasets, it is also encouraged that statistics information including geography, gender, ethnicity and other demographic information should be provided, for those datasets containing information about people.

\vspace{2mm}
\noindent\textbf{Intersectional Fairness}\, 
The investigation of intersectional fairness, i.e., combination of multiple sensitive attributes, is relatively lacking in current research~\cite{bose2019compositional,creager2019flexibly}.
Take bias mitigation for example, current work generally focus on one kind of bias. Although this may increase model fairness in terms of a specific bias, it is highly possible that the model is still biased from the intersectional perspective.
For instance, 
a gender-debiased DNN could be fair to women, while exhibiting discrimination towards a subdivision group, e.g., African American women or women over the age of 60.
Similarly for DNN based job recruiting tool,
even if the debiased model is free of gender bias, it is hard to guarantee that the model is not biased towards other protected attributes, e.g., race, age. 
More work is needed to figure out methods which are effective 
for identification and mitigation of intersectional biases.

\vspace{2mm}
\noindent\textbf{Fairness and Utility Trade-off}\, 
The removal of bias could possibly hurt the model's ability for main prediction task.
For instance, adversarial training could increase fairness. However, it could
compromise overall prediction accuracy, especially the accuracy for non-protected groups. Thus this might undermine the principle of beneficence. 
It remains a challenge to simultaneously reduce unintentional bias and maintain satisfactory model prediction performance.

\vspace{2mm}
\noindent\textbf{Formalization of Fairness}\, 
As the field of fairness machine learning is evolving quickly, there is still no consensus about the measurements of fairness. 
In certain cases, some measurements could be conflicting with others. 
A model may be fair in terms of one metric, but may lead to other sorts of unfairness.
For instance, a loan approval tool may satisfy demographic parity measurement, while violating equality of opportunity measurement.
There is no silver bullet, and each application domain calls for fairness measurements which meet its specific requirements~\cite{gajane2017formalizing}.

\subsection{Fairness in Large-scale Training}
Large-scale training is employed in some domains to boost model performance.
Take NLP domain for instance, current paradigm is to pre-train language models (e.g., BERT~\cite{devlin2018bert} and XLNet~\cite{yang2019xlnet}) on large-scale text corpus, which will be further fine-tuned on downstream tasks such as machine translation. These powerful language models could capture biases and propagate them to other tasks. Since these models need to be trained on corpus with billion-scale words and are typically trained for days, bias mitigation either through preprocessing or training regularization remains a challenge and more research is needed in this direction. 

\section{Conclusions}
With increasing adoption of DNNs in high-stake real world 
applications, e.g., job hunting, criminal justice and loan approval,  
their undesirable algorithmic 
unfairness problem has attracted much attention recently.
We give an overview of recent DNN bias detection and mitigation techniques from the computational perspective, with a particular focus on interpretability. 
The model bias to some extent exposes biases present in our society. To really benefit our society, DNN models are supposed to reduce these biases instead of amplifying biases. 
In future, endeavor from different disciplines, including computer science, statistics, cognitive science, should be joined together to eliminate disparity and promote fairness. In this way, DNN systems could be readily applied for fairness sensitive applications and
really improve benefits of our society. 



%




\ifCLASSOPTIONcaptionsoff
  \newpage
\fi



\bibliographystyle{IEEEtran}
\bibliography{fairness}
\end{document}